\documentclass[runningheads]{llncs}
\usepackage{graphicx}
\usepackage{diagbox}
\begin{document}
\title{Non-negative representation based discriminative dictionary learning for face recognition}
%
%

\author{Zhe~Chen\inst{1}
\and Xiao-Jun~Wu\inst{1}$^*$ \and Josef~Kittler\inst{2}
}

\authorrunning{Zhe~Chen et. al.}

\institute{Jiangsu Provincial Engineering Laboratory of Pattern Recognition and Computational Intelligence, School of IoT Engineering, Jiangnan University, Wuxi 214122, China \\
\and CVSSP, University of Surrey, GU2 7XH, Guildford, UK\\
}

\maketitle              
\begin{abstract}
In this paper, we propose a non-negative representation based discriminative dictionary learning algorithm (NRDL) for multicategory face classification. In contrast to traditional dictionary learning methods, NRDL investigates the use of non-negative representation (NR), which contributes to learning discriminative dictionary atoms. In order to make the learned dictionary more suitable for classification, NRDL seamlessly incorporates non-negative representation constraint, discriminative dictionary learning and linear classifier training into a unified model. Specifically, NRDL introduces a positive constraint on representation matrix to find distinct atoms from heterogeneous training samples, which results in sparse and discriminative representation. Moreover, a discriminative dictionary encouraging function is proposed to enhance the uniqueness of class-specific sub-dictionaries. Meanwhile, an inter-class incoherence constraint and a compact graph based regularization term are constructed to respectively improve the discriminability of learned classifier. Experimental results on several benchmark face data sets verify the advantages of our NRDL algorithm over the state-of-the-art dictionary learning methods.

\end{abstract}
\section{Introduction}

Due to the non-repeatability and uniqueness of human faces, face recognition has been the hottest topic in object classification applications ~\cite{3,4,5,6}. Over the past few years, sparse representation theory has been deeply studied in the field of face recognition. The representative one is sparse representation-based classification (SRC) ~\cite{7} algorithm. SRC aims at linearly representing an input data with a few atoms from given training samples. Zhang et al. ~\cite{11} considered that the inter-class collaboration plays a more important role than sparsity. Therefore, they proposed a collaborative representation-based classification (CRC) algorithm by imposing an $L_2$-norm constraint on representation coefficients which can achieve competitive classification accuracy in less time than $L_1$-norm based methods. Besides, Xu et al. ~\cite{13} proposed a new discriminative sparse representation method, which suggests that the discriminative representation can be obtained by suppressing the relevance of inter-class reconstructions. More recently, Xu et al. ~\cite{14} found that non-negative representation can strengthen the representation ability of homogeneous samples while weakening the negative effects caused by heterogeneous samples. Note that above representation based algorithms can only work well when the samples are collected under well-controlled conditions. However, in reality, face images often include severe illumination, pose, expression and occlusion changes that can destroy the subspace structure of data. So it is necessary to learn some distinct atoms which are beneficial for representation. Recently, as a major branch of sparse representation, dictionary learning has attracted extensive interests in face recognition field. The main idea of dictionary learning is to extract competitive and discriminative features from original training samples, while removing the useless information that is bad for reconstruction. 

According to whether the label information of training samples is used, dictionary learning algorithms are usually divided into two categories: supervised or unsupervised.  Xu et al. ~\cite{18} proposed a sample-diversity and representation-effectiveness based robust dictionary learning algorithm (SDRERDL) by taking advantages of the mirroring samples to address the small-sample-size problem. Different from unsupervised situation, supervised dictionary learning algorithms can extract more discriminative features from data with the use of label information.  Zhang et al. ~\cite{19} and Jiang et al. ~\cite{20} proposed a discriminative KSVD (D-KSVD) algorithm and a label-consistent K-SVD (LC-KSVD) algorithm, respectively. Yang et al. ~\cite{21} proposed a famous Fisher discrimination criterion to encourage the discriminability of learned dictionary by simultaneously minimizing within-class scatter and maximizing inter-class scatter of representation.  Recently, based on the observation that locality of data may be more significant than sparsity, Li et al. ~\cite{24} proposed a locality-constrained and label embedding dictionary learning algorithm (LCLE), which considers the locality and label information of samples together during the dictionary learning process.

It is worth noting that aforementioned dictionary learning algorithms exist a common problem: the representation coefficients for training samples may include negative values. According to ~\cite{14}, since there is no non-negative constraint on representation, a query sample will be represented by both heterogeneous and homogeneous samples. As a result, the obtained coefficients have both negative and non-negative values, which brings about a difficult physical interpretation. Based on the assumption that a query sample should be approximated by homogeneous samples as much as possible, in this paper, we propose a non-negative representation based discriminative dictionary learning algorithm (NRDL) for face recognition. Specifically, the contributions
of the proposed NRDL algorithm are presented as follows.

(1) By restricting the coding values to be non-negative, the NRDL model will select some useful features from heterogeneous samples to form dictionary atoms, which naturally leads to discriminability and sparsity on representation.  

(2) Due to the combination of dictionary learning and multi-class classifier learning, the learned dictionary is efficient for reconstruction and classification simultaneously.

(3) NRDL minimizes the inter-class reconstruction to encourage the discriminability of learned dictionary instead of directly forcing the representation to be a block-diagonal structure which may result in overfitting problem. 

(4) For better classification, NRDL uses an inter-class incoherence term and a compact graph structure to improve the robustness of learned linear classifier.

\section{Non-negative representation for classification (NRC)}

Among previously mentioned representation based classification methods ~\cite{7}~\cite{11}, there still exists controversy between sparsity and collaborative mechanism, that is to say which mechanism is more significant when using training samples to approximate a query sample? Nevertheless, Xu et al. ~\cite{14} thought it is meaningless to determine an eventual winner. Despite the successful applications of sparsity and cooperativity, they proposed a simple but efficient non-negative representation-based classification algorithm (NRC) by introducing a non-negative constraint on coding coefficients instead of using sparse constraints (e.g., $l_1$ or $l_2$ norm). Assuming that we have $N$ training samples from $C$ classes, denoted by $X=[X_1, X_2, ... , X_C]\in R^{d \times N}$, where $X_i\in R^{d \times n}(n=N/C)$ denotes the training samples of $i$th class . $d$ is the sample dimensionality. Given a query sample $y\in R^d$, the model of NRC can be formulated as:
$$\min_{\theta} \|y-X\theta\|_2^2 \quad  s.t. \quad \theta>0 \eqno{(1)}$$
where $\theta \in R^N$ is the coding vector. 
NRC argues that the non-negative coefficients over the homogeneous samples are crucial to classify $y$. By restricting the values of coding vector $\theta$ to be non-negative, the contributions of homogeneous samples can be enlarged, meanwhile, eliminating the negative effects caused by heterogeneous samples. Instead of using original training samples as the dictionary, we extend the non-negativity theory of representation to dictionary learning method, so that the learned dictionary atoms are more high-quality and homogeneous. In addition, we also investigate how to enhance the discriminability and compact of learned dictionary and classifier, respectively.

\section{Non-negative representation based dictionary learning algorithm (NRDL)}
In this section, we present a novel non-negative representation-based dictionary learning algorithm (NRDL), in which dictionary learning and linear classifier learning are incorporated into a joint framework. In NRDL, we want to learn a discriminative and reconstructive dictionary by leveraging the label information of training samples under the condition that the coding values are non-negative. Because representation of samples can also be regarded as feature for classification, the classification error term is also included in our learning model to make the learned dictionary efficient for classification as well. Thus, the optimization framework of NRDL can be defined as follows
\setcounter{equation}{1} 
\begin{eqnarray}
<D,S,W>=arg\min_{D,S,W}h(D,S)+ g(H,W,S)\ \ s.t.  \   S\geq0
\end{eqnarray}
where $h(D)$ is the discriminative dictionary encouraging function and $g(H,W,S)$ is the robust linear classifier learning function. $H=[h_1, h_2, ... , h_n]\in R^{C \times N}$ are the class labels of training samples $X$. Specifically, $h_i=[0,0,...,1,...,0,0]\in R^C$ is a label vector of $i$th training sample $x_i$ where the position of element '1' indicates the class of $x_i$. $D=[D_1, D_2, ... , D_C]\in R^{d \times K}$ is the learned dictionary and $S=[S_1, S_2, ... , S_C]\in R^{K \times N}$ are the sparse codes of training samples, where $D_i$ and $S_i$ $(i=1,2,...,C)$ denote the sub-dictionary and sub-representation corresponding to class $i$, respectively. $S\geq0$ represents the non-negative constraint on coding matrix $S$. $W\in R^{C\times K}$ is the projection matrix of linear classifier. $K$ is the number of dictionary atoms.  Now, we will detail the objective functions of $h(D,S)$ and $g(H,W,S)$. 

\subsection{Non-negative discriminative dictionary encouraging function}
Obviously, for the $l$th class training samples $X_l$, we have $X_l=DS_l$. Supposing $S_{ii}$ is the coding coefficients of $X_i$ over sub-dictionary $D_i$, note that $S_{ll}$ should be capable of well reconstructing $X_i$ with $D_i$. so it is reasonable to assume that $X_i\approx D_iS_{ii}$. What's more, it is expected that $X_i$ can only be well reconstructed by $S_{ii}$, rather than by inter-class representation $S_{ji}(j=1,2,...,C, j\neq i)$. In order to learn a discriminative and reconstructive dictionary, Zhang et al.[23] argued the representation matrix to present an ideal '0-1' block-diagonal structure $Q=[q_1,q_2,...,q_n]\in R^{K\times n}$ to capture more structured information from data. Let sample $x_i$ belongs to class $l$, then the coding coefficients in $q_i$ for $D_l$ are all 1s, while the remaining elements are all 0s. However, the samples from the same class often have different coding coefficients because of the diversity of samples. To address above problems, we propose the following non-negative discriminative dictionary encouraging function for our NRDL method     
\setcounter{equation}{2} 
\begin{eqnarray}
h(D,S)=\|X-DS\|{^2_F}+\lambda\|D(A\odot S)\|^2_F \ s.t.  \ S\geq0 
\end{eqnarray}
where $\lambda$ is a positive parameter. $A=O-Q$, where $O=\textbf{1}_K\textbf{1}_N^T\in R^{K\times N}$ denotes all 1s matrix. $Q=[q_1, q_2,...,q_N]\in R^{K\times N}$ denotes the '0-1' block-diagonal matrix, where $q
_i$ has the form of $[0,1,1,1,...]^T$. If sample $x_i$ belongs to class $L$, then the coding in $q_i$ over $D_L$ are all 1s, while all the others are 0s. $\|X-DS\|{^2_F}$ is the reconstruction error term. We can see that $A\odot S$ is actually the off-block-diagonal components of representation matrix $S$. Based on the observation that the samples corresponding to a certain class should only be well represented by the sub-dictionary from the same class, hence it is natural to assume the off-block-diagonal reconstruction with dictionary $D$, i.e.,$(D(A\odot S))$, should be minimized as much as possible, thus enhancing the uniqueness of class-specific sub-dictionaries.  

\subsection{Robust linear classifier learning function}
Similar to ~\cite{19} and ~\cite{20}, we consider to introduce the classification error term in our NRDL framework to learn a dictionary which is also suitable for classification. Besides, we also propose another two regularization terms to improve the discriminability of learned classifier. The proposed robust classifier learning function can be formulated as
\setcounter{equation}{3} 
\begin{eqnarray}
&g(H,W,S)= \alpha \|H-WS\|_F^2+&  \nonumber \\ 
&\beta\sum_{i=1}^C\sum_{j=1,j\neq i}^C\|(WS_i)^T(WS_j)\|_F^2+\gamma\sum_{v=1}^N\sum_{u=1}^N M_{uv}\|WS_v-WS_u\|_2^2&
\end{eqnarray}
where $\alpha$, $\beta$ and $\gamma$ control the weights of regularization terms. $\|H-WS\|_F^2$ represents the reconstruction error term. Given a sample $x_i\in X$, $f(x_i;W)=Wx_i$ is the linear predictive classifier. The second term is the inter-class classification projections incoherence promoting term. $WS_i$ and $WS_j$ denote the projection results of $i$th class and $j$th class, respectively. By minimizing the second term, the independency of inter-class classification projections will be encouraged, hence improving the discriminability of classifier. Although the classifier is discriminative, the intra-class compact is also crucial to robust classifier learning. Inspired by the idea of manifold learning, the intra-class compact graph is built in the transformed space. In our model the transformed space is generated by projecting the coding vector of samples and each graph node, i.e., $WS_u$, represents the classification projection of coding vector $S_u$ which corresponding to sample $x_u$. Because the label vectors of the samples from the same classes are the same, we think that their projections should be maintained close together.  In the third term of formula (4), $M_{uv}$ denotes the weight between two graph nodes corresponding to two different samples $x_u$ and $x_v$. Referring to LPP ~\cite{33}, we use the sample similarity to define the weight of compact graph. In our paper, the similarity between samples is calculated under following unsupervised form
$$M_{uv}=\frac 1 {1+\|x_u-x_v\|^2_2} \eqno{(5)}$$

From formula (5), we can find if samples $x_u$ and $x_v$ have the same labels, although their distance $\|x_u-x_v\|^2_2$ is intrinsically small, their similarity (graph weight) $M_{uv}$ is big. Inversely,  if $x_u$ and $x_v$ are from different classes, their distance is intrinsically big, but their similarity is small. Hence,  minimizing the third term will simultaneously enhance the intra-class compactness and the inter-class discriminability of classifier. According to (25), we can rewrite function (4) as
\setcounter{equation}{5} 
\begin{eqnarray}
&g(H,W,S)= \alpha \|H-WS\|_F^2+&  \nonumber \\ 
&\beta\sum_{i=1}^C\sum_{j=1,j\neq i}^C\|(WS_i)^T(WS_j)\|_F^2+\gamma tr(WSLS^TW^T)&
\end{eqnarray} 
where $L=Z-M$ is the graph laplacian matrix. $Z$ is a diagonal matrix and $Z_{ii}=\sum_{j=1}^{N} M_{ij}$.
By combining the non-negative discriminative dictionary encouraging function (3) and the robust linear classifier learning function (6), the final formulation of the proposed NRDL algorithm is
\setcounter{equation}{6} 
\begin{eqnarray}
<D,S,W>=arg\min_{D,S,W}\|X-DS\|{^2_F}+ \lambda\|D(A\odot S)\|^2_F+ \nonumber \\
\alpha \|H-WS\|_F^2+ \beta\sum_{i=1}^C\sum_{j=1,j\neq i}^C\|(WS_i)^T(WS_j)\|_F^2+\nonumber \\ 
\gamma tr(WSLS^TW^T) \quad s.t.  \ S\geq0 
\end{eqnarray}

\section{Solving the optimization problem of NRDL}
We can find it is impossible to directly solve problem (7) because the variables, i.e., $D$, $S$, $W$, are interactional. In this section, we use the alternating direction method of multipliers (ADMM) ~\cite{26} to update variables one by one, which means when updating one variable, all the others should be fixed. We first introduce two auxiliary variables to make the problem separable. Therefore, problem (7) can be rewritten as
\setcounter{equation}{7} 
\begin{eqnarray}
&<D,S,W,P,J>=arg\min_{D,S,W,P,J}\|X-DS\|{^2_F}+ \lambda\|D(A\odot S)\|^2_F+& \nonumber \\
&\alpha \|H-WS\|_F^2+ \beta\sum_{i=1}^C\sum_{j=1,j\neq i}^C\|P_i^T P_j\|_F^2+\gamma tr(JLJ^T)& \nonumber \\
&\quad s.t. \ WS=P, WS=J,  S\geq0 &
\end{eqnarray}
Then, the augmented Lagrangian function $L_\mu$ of problem (8) is defined as
\setcounter{equation}{8} 
\begin{eqnarray}
&L_\mu(D,S,W,P,J,C_1,C_2)=\|X-DS\|{^2_F}+ \lambda\|D(A\odot S)\|^2_F+& \nonumber \\
&\alpha \|H-WS\|_F^2+ \beta\sum_{i=1}^C\sum_{j=1,j\neq i}^C\|P_i^T P_j\|_F^2+\gamma tr(JLJ^T)+& \nonumber \\
&\mu\|WS-P+\frac{C_1}{\mu}\|_F^2+\mu\|WS-J+\frac{C_2}{\mu}\|_F^2-\frac{1}{\mu}(\|C_1\|_F^2+\|C_2\|_F^2)& \nonumber \\
&s.t. \ S\geq0 &
\end{eqnarray}
where $C_1$ and $C_2$ are the Lagrangian multipliers, and $\mu>0$ is the penalty parameter. Next starting the iterations:

\noindent\textbf{Update S:} By fixing variables $J$, $W$, $P$, $D$, $C_1$ and $C_2$, we update $S$ as 
\setcounter{equation}{9} 
\begin{eqnarray}
S^{k+1}=arg\min_S\|X-DS\|{^2_F}+ \lambda\|D(A\odot S)\|^2_F+& \nonumber \\
\alpha \|H-WS\|_F^2+\mu\|WS-P+\frac{C_1}{\mu}\|_F^2+\mu\|WS-J+\frac{C_2}{\mu}\|_F^2
&s.t. \ S\geq0 &
\end{eqnarray}

Because $A=O-Q$, $D(A\odot S)=D[(O-Q)\odot S]=D(S-Q\odot S)=DS-D(Q\odot S)$. Let $R=D(Q\odot S^k)$, $\|D(A\odot S)\|^2_F$ becomes $\|DS-R\|^2_F$. By making the derivation of (10) with respect to $S$, the optimal $S$ can be calculated as
\setcounter{equation}{10} 
\begin{eqnarray}
&S^{k+1}=[(\lambda+1)D^k(D^k)^T+(\alpha+2\mu)(W^k)^TW^k]^{-1}[(D^k)^T(X+\lambda R)+& \nonumber \\ 
&(W^k)^T(\alpha H+\mu P^k - C_1^k+\mu J^k-C_2^k)]&
\end{eqnarray}
then in each iteration, we change the negative elements in $S$ to be 0, thus generating non-negative representation.

\noindent\textbf{Update W, P, J and D:} 
$$W^{k+1}=(\alpha H+\mu P_k - C_1^k+\mu J^k - C_2^k)(S^{k+1})^T[(\alpha+2\mu)S^{k+1}(S^{k+1})^T]^{-1} \eqno{(12)}$$

$$P_i^{k+1}=(\sum_{j=1, j\neq i}^C P_j^k (P_j^k)^T+\mu I)^{-1}(\mu W^{k+1}S_i^{k+1} + C_{1i}^k) \eqno{(13)}$$

$$J^{k+1}=(\mu W^{k+1}S^{k+1} + C_2^k)(\gamma L + \mu I)^{-1} \eqno{(14)}$$

$$D^{k+1}=XS^{k+1}[S^{k+1}(S^{k+1})^T+\lambda MM^T]^{-1} \eqno{(15)}$$

\noindent\textbf{Update Lagrangian multipliers $C_1$ and $C_2$:} 
$$C_1^{k+1}=C_1^k+\mu(W^{k+1}S^{k+1}-P^{k+1}) \eqno{(16)}$$
$$C_2^{k+1}=C_2^k+\mu(W^{k+1}S^{k+1}-J^{k+1}) \eqno{(17)}$$

\noindent\textbf{Check the convergence}:
$$if \max\{\|W^{k+1}S^{k+1}-P^{k+1}\|_{\infty}, \|W^{k+1}S^{k+1}-J^{k+1}\|_{\infty}\}\leq tol. \eqno{(18)}$$
then stop the iterations.

\section{Classification}
After the discriminative dictionary $D=[D_1,D_2,...,D_C]\in R^{d\times K}$ and robust linear classifier $W\in R^{C\times K}$ are obtained by NRDL, given a test sample $x_{new}\in R^{d}$, we first calculate its coding vector over the learned dictionary $D$. Because the dictionary and classifier are both learned with non-negative representation, here we use the NRC ~\cite{14} model to solve the coding vector of $x_{new}$ 
$$\min_{\eta} \|x_{new}-D\eta\|_2^2 \quad  s.t. \quad \eta>0 \eqno{(19)}$$

The used NR model is primarily the non-negative least squares problem, which does not have a closed-form solution. Referring to literature ~\cite{14}, we also utilize ADMM ~\cite{26} to solve the NR model. Once we get the optimal coding vector $\hat\eta$, the classification of sample $x_{new}$ is similar to algorithms ~\cite{24} (See Algorithm 1 for details). 

\begin{table}[!ht]
\rule[0.05cm]{12.5cm}{2pt}
\leftline {\textbf {Algorithm 1.} The classification approach based on NR model}\\
\rule[0.1cm]{12.5cm}{2pt}
\textbf{Input:} Learned dictionary $D$ and classification projection matrix $W$, outside test sample $x_{new}$
 \begin{enumerate}
\item Normalize $x_{new}$ to have unit $l_2$ norm;
\item Code $x_{new}$ over dictionary $D$ via solving problem (19):
\item Calculate the classifier projection of coding vector $\hat\eta$: $f=W\hat\eta$;
\end{enumerate}

\textbf{Output:} $Label(x_{new})=arg\max_{i}\{f_i\}$, where $f_i$ represents the $i$th entry of $f$.\\
\rule[0.2cm]{12.5cm}{2pt}
\end{table}

\section{Experiments}
In this section, some experiments were performed on five benchmark face databases: ORL ~\cite{27}, GT ~\cite{28}, CMU PIE ~\cite{29}, Extended Yale B ~\cite{30}, and Labeled Faces in the wild (LFW) ~\cite{31}. To evaluate the performance of proposed NRDL algorithm, we compared it with the LLC ~\cite{18}, SRC ~\cite{7}, LRC ~\cite{32},  D-KSVD ~\cite{19}, LC-KSVD2 ~\cite{20}, FDDL ~\cite{21}, SDRERDL ~\cite{18}, and LCLE ~\cite{24} algorithms. LLC solves the coding coefficients by utilizing the approximate LLC method. LLC, LRC, and SRC have no dictionary learning method and directly use the original training samples as their dictionary. All the compared dictionary learning algorithms used the same classification approach (linear classification). We used the LC-KSVD algorithm to initialize the dictionary. Following ~\cite{20}, sparsity factor $\xi$=30 was used in the K-SVD, D-KSVD, LC-KSVD2, SDRERDL, and LCLE algorithms. For the sake of fairness, the number of local bases was identical to $\xi$ in the LLC algorithm. Besides, in order to verify the effects of non-negative representation, we also tested the performance of our algorithm without non-negative constraint on representation (NRDL-test), then the CRC model ~\cite{11} was used to solve the coding coefficients of test samples in NRDL-test. As shown in ~\cite{18} and ~\cite{24}, the average recognition rates of almost all algorithms increased with the increase in the number of dictionary atoms. This is mainly because the reconstruction and discriminative ability of the dictionary improve with the increase of the number of atoms. Thus, in this paper, we tested all the dictionary learning algorithms with setting the number of dictionary atoms to the size of original training samples. The brief description of these datasets are shown in Tabel 1. 

\begin{table}[]
\setlength{\tabcolsep}{3.8pt}
\renewcommand{\arraystretch}{1.5}
\caption{Brief description of the used five datasets.}
\centering
\begin{tabular}{|c|c|c|c|c|c|}
\hline
\diagbox{Dataset}{Info.} & Classes & Features & Total Num. & Train Num. & K \\
\hline
ORL & 40 & 2576 & 400 & 240 & 240 \\
\hline
GT & 50 & 2000 & 750 & 250 & 250\\
\hline
CMU PIE & 68 & 1024 & 11554 & 680 & 680\\
\hline
Extended Yale B & 38 & 1024 & 2414 & 760 & 760\\
\hline
LFW & 86 & 1024 & 1251 & 688 & 688\\
\hline
\end{tabular}
\end{table} 

\subsection{Experimental results on five face databases}

The average recognition rates on five face databases are reported in Table 2. The sign $\pm$ represents the standard deviation of ten times results. As shown in Table 2, the proposed NRDL algorithms achieves almost the best recognition rates in comparison with all of the compared algorithms on different databases. This
indicates that our NRDL algorithm is capable of effectively learning a discriminative dictionary from original samples under the condition of enforcing representation to be non-negative, and the learned dictionary is simultaneously robust for classification. Moreover, we can see that the recognition performance of NRDL is better than NRDL-test in all databases, especially in GT, CMU PIE and LFW databases. This is mainly because that the non-negative representation is beneficial for learning distinct features and eliminating useless information from heterogeneous data.

\makeatletter\def\@captype{table}\makeatother

\renewcommand\arraystretch{1.5}
\begin{table}
\caption{\label{tab1}Average recognition rates (\%) of different algorithms on five databases }
\begin{tabular}{|c|c|c|c|c|c|}
\hline
\diagbox{Algorithms}{Databases} & ORL & GT & CMU PIE & Extended Yale B & LFW  \\ 
\hline 
LLC  & 93.1 & 60.0 & 53.7$\pm$0.016 & 88.9$\pm$0.010 & 34.8$\pm$0.011  \\
\hline
LRC  & \textbf{94.4} & 59.4 & 61.6$\pm$0.021 & 92.4$\pm$0.008 & 37.1$\pm$0.014   \\
\hline
SRC  & \textbf{94.4} & \textbf{63.8} & 72.1$\pm$0.008 & 95.3$\pm$0.005 & 38.1$\pm$0.011   \\
\hline
D-KSVD  & 93.8 & 56.6 & 71.9$\pm$0.008 & 83.0$\pm$0.026 & 33.4$\pm$0.016    \\
\hline
LC-KSVD2  & 92.5 & 56.0 & 72.3$\pm$ 0.009 & 92.7$\pm$0.008 & 32.2$\pm$0.012    \\
\hline
FDDL  & 93.8 & 62.4 & 70.6$\pm$0.020 & 93.0$\pm$0.008 & 41.7$\pm$0.016  \\
\hline
SDRERDL  & 93.8 & 57.6 & 77.0$\pm 0.006$ & 96.0$\pm$0.004 & 37.3$\pm$0.013  \\
\hline
LCLE  & 91.9 & 58.6 & 75.6$\pm$0.009 & 95.8$\pm$0.005 & 38.8$\pm$0.009 \\
\hline
\textbf{NRDL-test} & 92.5 & 56.6 & 75.0$\pm$0.009 & 93.2$\pm$0.009 & 35.0$\pm$0.010 \\
\hline
\textbf{NRDL(ours)} & \textbf{94.4} & \textbf{63.6} & \textbf{81.0$\pm$ 0.009} & \textbf{96.3$\pm$ 0.004} &  \textbf{43.1$\pm$ 0.014} \\
\hline
\end{tabular}
\end{table}

\subsection{Experimental results with 'Salt and Pepper' noise}
To investigate the robustness of NRDL algorithm, we tested the performance of proposed NRDL algorithm and some relatively new dictionary algorithms, i.e., FDDL, SDRERDL and LCLE, by adding artificial noise in samples. In our experiments, we obtain contaminated images by using the Matlab function "imnoise" to impose 'Salt\&Pepper' noise on all the original face images. In order to verify the performance with different degrees of contaminations, the density of noise (i.e., the third parameter of "imnoise" function) are set to 0.01, 0.02, and 0.03, respectively. The way of selecting samples of different databases are the same as the previous section. All experiments ran ten times and the rate was averaged. The average recognition rates of different algorithms are shown in Fig.1. Fig.1 shows that the average recognition rates of the comparison algorithms and the proposed NRDL algorithm all decrease with the increase in the contamination degree of images. This is mainly because the 'Salt\&Pepper' noise can destroy the subspace structure of data so that the learned features are not sufficient and distinct. Moreover, we can see the accuracy of our NRDL algorithm outperforms all the comparison algorithms on four databases, which demonstrates that NRDL is indeed more robust to artificial noise.

\begin{figure}[h]
\centering
\includegraphics[scale=0.5]{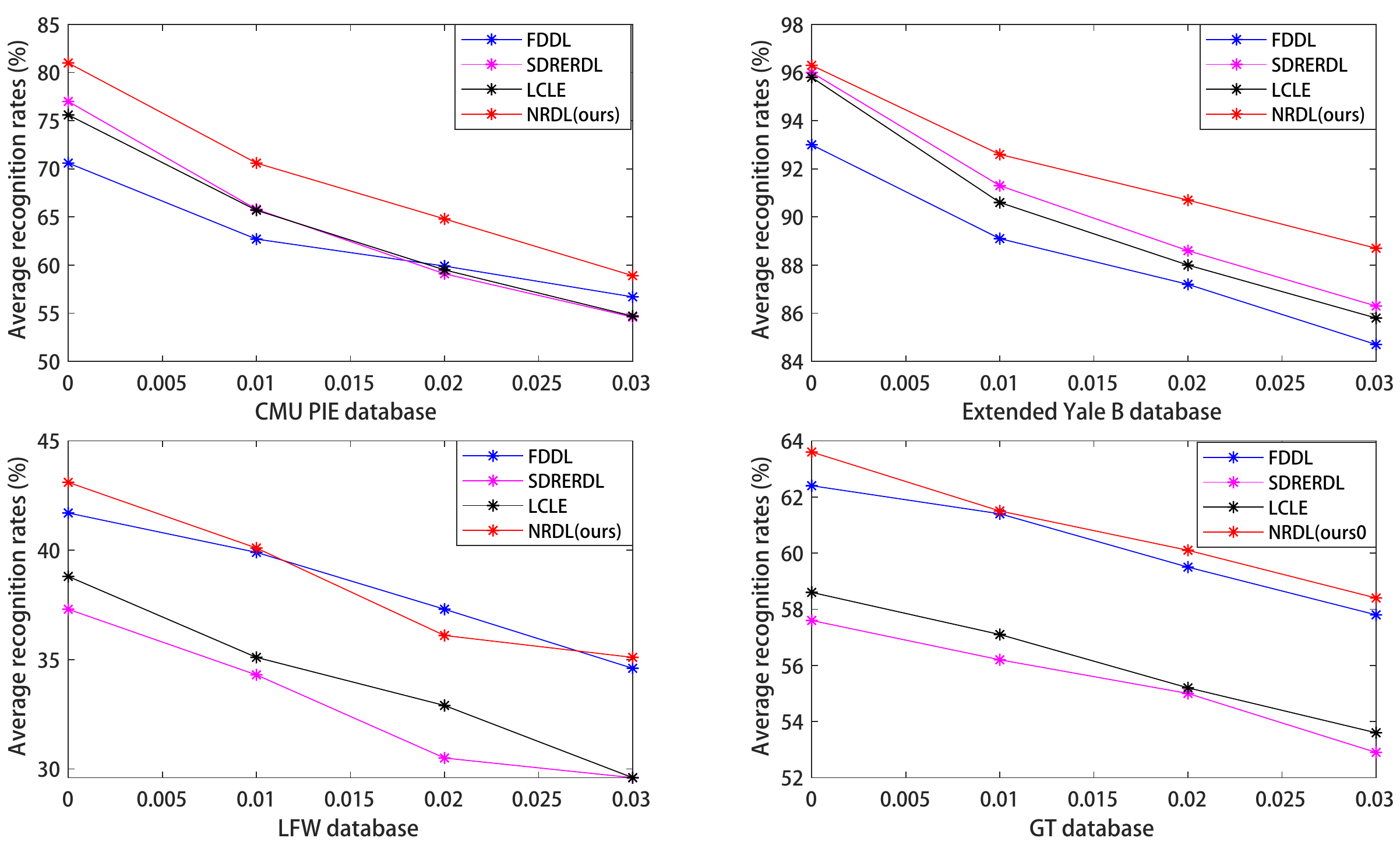}
\caption{Average recognition rates under different degrees of 'Salt\&Pepper' noise contamination}
\end{figure}

\subsection{Convergence validation}
The convergence of the proposed NRDL algorithm on four databases is illustrated in Fig.2. As
expected, the convergence of the objective function (7) is very fast. 

\begin{figure}[h]
\centering
\includegraphics[scale=0.65]{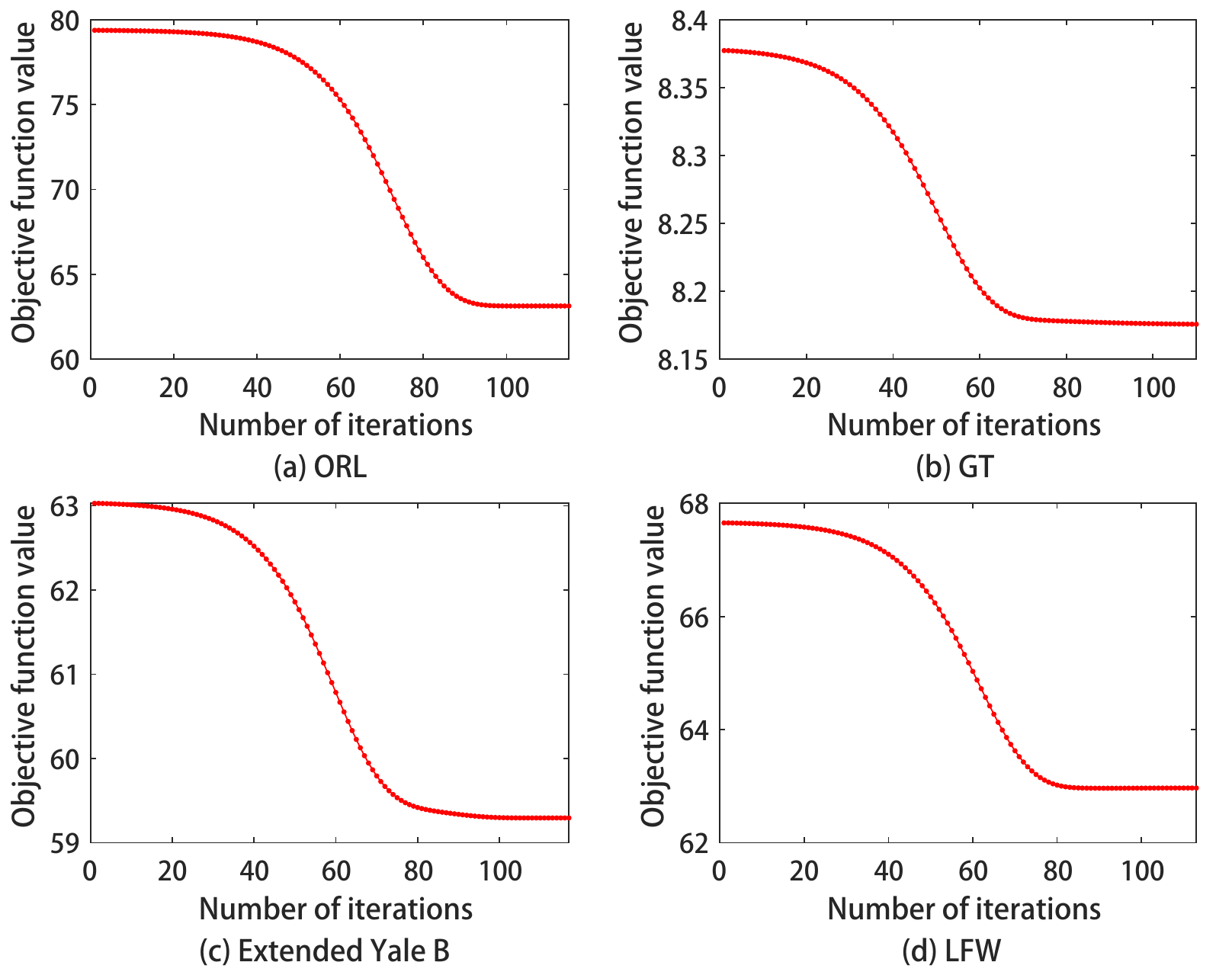}
\caption{Objective function value versus the number of iterations of the proposed NRDL algorithm on four databases.}
\end{figure}

\section{Conclusion}
In this paper, a non-negative representation based discriminative dictionary learning algorithm (NRDL) is proposed for multi-class face classification. Different from other dictionary learning methods, NRDL considers to learn a more discriminative and reconstructive dictionary with non-negative representation constraint. Specifically, NRDL is designed to incorporate non-negative representation learning, discriminative dictionary learning and robust linear classifier learning into a unified framework. As a result, the learned dictionary is effective for classification. Experimental results on five face databases indicate that the NRDL algorithm is superior to the nine state-of-the-art sparse coding and dictionary learning algorithms.

%
%
%
%

\end{document}